\documentclass[a4paper,twocolumn,11pt,unpublished]{quantumarticle}
\pdfoutput=1

\usepackage{microtype}
\usepackage{graphicx}
\usepackage{subfigure}
\usepackage{booktabs}
\usepackage{amssymb}
\usepackage{mathtools}
\usepackage{amsthm}
\usepackage[utf8]{inputenc}

\usepackage{amsmath}
\usepackage{booktabs}
\usepackage{multirow}

\usepackage{bm}
\usepackage{caption}
\usepackage{subcaption}

\usepackage[
  style=numeric-comp,
  sorting=none,
  defernumbers=false,
  sortcites=false,
  backend=biber,
  doi=false,
  url=false,
  hyperref=auto,
  giveninits=true,
  uniquename=false,
  maxbibnames=6,
  minbibnames=6
]{biblatex}
\usepackage{hyperref}

\DeclareFieldFormat[article]{title}{#1}
\DeclareFieldFormat[inproceedings]{title}{#1}
\DeclareFieldFormat[incollection]{title}{#1}
\DeclareFieldFormat[book]{title}{#1}

\renewbibmacro{in:}{}

\DeclareFieldFormat{pages}{#1}
\DeclareFieldFormat{pagetotal}{#1}

\renewbibmacro*{title}{%
  \printtext[title]{%
    \printfield[titlecase]{title}%
    \setunit{\subtitlepunct}%
    \printfield[titlecase]{subtitle}%
  }%
}

\newbibmacro*{journalline}{%
  \usebibmacro{journal}% journal name
  \setunit*{\addspace}%
  \usebibmacro{volume+number+eid}% e.g. 11 6 or 11(6)
  \setunit{\addcomma\space}%
  \printfield{pages}%
  \setunit{\addspace}%
  \printtext[parens]{\printfield{year}}%
}

\renewbibmacro*{journal+issuetitle}{%
  \iffieldundef{doi}
    {%
      \iffieldundef{url}
        {\usebibmacro{journalline}}%
        {\href{\thefield{url}}{\usebibmacro{journalline}}}%
    }%
    {\href{https://doi.org/\thefield{doi}}{\usebibmacro{journalline}}}%
  \newunit
}

\renewbibmacro*{note+pages}{}

\renewbibmacro*{date+extradate}{}

\renewbibmacro*{volume+number+eid}{%
  \printfield{volume}%
  \iffieldundef{number}{}{\setunit*{\addnbthinspace}\printfield{number}}%
  \iffieldundef{eid}{}{\setunit*{\addnbthinspace}\printfield{eid}}%
}

\newbibmacro*{proceedingsline}{%
  \iffieldundef{booktitle}
    {\printfield[titlecase]{maintitle}}
    {\printfield[titlecase]{booktitle}}%
  \iffieldundef{booksubtitle}{}{%
    \setunit{\subtitlepunct}\printfield[titlecase]{booksubtitle}}%
  \iffieldundef{pages}{}{\setunit{\addcomma\space}\printfield{pages}}%
  \setunit{\addspace}\printtext[parens]{\printfield{year}}%
}

\renewbibmacro*{maintitle+booktitle}{%
  \iffieldundef{doi}
    {%
      \iffieldundef{url}
        {\usebibmacro{proceedingsline}}%
        {\href{\thefield{url}}{\usebibmacro{proceedingsline}}}%
    }%
    {\href{https://doi.org/\thefield{doi}}{\usebibmacro{proceedingsline}}}%
  \newunit
}

\renewbibmacro*{event+venue+date}{}   % we'll keep it minimal; remove if you want venues
\renewbibmacro*{note+pages}{}         % already printed inside proceedingsline

\newbibmacro*{bookpubline}{%
  \printlist{publisher}%
  \iffieldundef{location}{}{\setunit*{\addcomma\space}\printlist{location}}%
  \setunit{\addspace}\printtext[parens]{\printfield{year}}%
}

\renewbibmacro*{publisher+location+date}{%
  \iffieldundef{doi}
    {%
      \iffieldundef{url}
        {\usebibmacro{bookpubline}}%
        {\href{\thefield{url}}{\usebibmacro{bookpubline}}}%
    }%
    {\href{https://doi.org/\thefield{doi}}{\usebibmacro{bookpubline}}}%
  \newunit
}

\renewbibmacro*{doi+eprint+url}{%
  \iftoggle{bbx:doi}{\printfield{doi}}{}%
  \newunit\newblock
  \iftoggle{bbx:eprint}{\usebibmacro{eprint}}{}%
  \newunit\newblock
  \iftoggle{bbx:url}{\usebibmacro{url+urldate}}{}%
}

\usepackage[svgnames]{xcolor}
\hypersetup{
    hidelinks,
    colorlinks=true,
    breaklinks=true,
    citecolor=MediumPurple,
    filecolor=LimeGreen,
    linkcolor=MediumBlue,
    urlcolor=MediumPurple
}

\DeclareUnicodeCharacter{2236}{\ensuremath{:}}          % ∶  (ratio)
\DeclareUnicodeCharacter{2192}{\ensuremath{\rightarrow}}% →  (arrow)
\DeclareUnicodeCharacter{2295}{\ensuremath{\oplus}}     % ⊕  (circled plus)
\DeclareUnicodeCharacter{27E9}{\ensuremath{\rangle}}    % ⟩  (ket bracket)
\DeclareUnicodeCharacter{2212}{-}                       % −  (real minus)
\DeclareUnicodeCharacter{2208}{\ensuremath{\in}}        % ∈  (element of)
\DeclareUnicodeCharacter{2013}{--}                      % –  (en dash, e.g. in Laguerre–Gaussian)
\DeclareUnicodeCharacter{2113}{\ensuremath{\ell}}       % ℓ  (script l)
\DeclareUnicodeCharacter{03C8}{\ensuremath{\psi}}       % ψ  (psi)
\DeclareUnicodeCharacter{03C6}{\ensuremath{\phi}}       % φ  (phi)
\DeclareUnicodeCharacter{2090}{\ensuremath{_{\mathrm{a}}}} % ₐ  (subscript a)
\DeclareUnicodeCharacter{2211}{\ensuremath{\sum}}       % ∑  (summation)
\DeclareUnicodeCharacter{2081}{\ensuremath{_{1}}}       % ₁  (subscript 1)
\DeclareUnicodeCharacter{2082}{\ensuremath{_{2}}}       % ₂  (subscript 2)

\usepackage{tikz}
\usepackage{fvextra} % loads fancyvrb and extends it
\usepackage[most]{tcolorbox}

\newenvironment{myverbatim}
  {%
    \VerbatimEnvironment
    \begin{tcolorbox}[
      enhanced,
      breakable,
      colback=gray!10,
      colframe=gray!50,
      boxrule=0.5pt,
      arc=2mm,
      outer arc=2mm,
      left=1ex, right=1ex,
      top=0.7ex, bottom=0.7ex,
      fontupper=\small\ttfamily % <-- ensure monospaced font
    ]%
    \begin{Verbatim}[
      breaklines=true,
      breakanywhere=true,
      breaksymbolleft={},
      breaksymbolindent=0pt,
      breakautoindent=false,
      breakindent=0pt
    ]%
  }
  {%
    \end{Verbatim}
    \end{tcolorbox}
  }

\usepackage{listings}

\usepackage{xspace} 
\newcommand{\PyTheus}{\textnormal{\textsc{PyTheus}}\xspace}
\let\pytheus\PyTheus  % keep your lowercase alias working

\newcommand{\impactfourcast}{\textsc{Impact4Cast}\xspace}

\newcommand{\scimuse}{\textsc{SciMuse}\xspace}

\newcommand{\AIMandel}{\textnormal{\textsc{AI-Mandel}}\xspace}

\newcommand{\mpl}{Max Planck Institute for the Science of Light, Erlangen, Germany}
\newcommand{\jena}{Institut für Festkörpertheorie und Optik, Friedrich-Schiller-Universität Jena, Jena, Germany}
\newcommand{\tuebingen}{Machine Learning in Science Cluster, Department of Computer Science, Faculty of Science, University of Tuebingen, Germany}

\begin{document}

\title{Towards autonomous quantum physics research using LLM agents with access to intelligent tools}

\author{Sören Arlt}
\email{soeren.arlt@uni-tuebingen.de}
\affiliation{\tuebingen}
\affiliation{\mpl}

\author{Xuemei Gu}
\email{xuemei.gu@uni-jena.de}
\affiliation{\jena}

\author{Mario Krenn}
\email{mario.krenn@uni-tuebingen.de}
\affiliation{\tuebingen}
\affiliation{\mpl}
\maketitle
\begin{abstract}
Artificial intelligence (AI) is used in numerous fields of science, yet the initial research questions and targets are still almost always provided by human researchers. AI-generated creative ideas in science are rare and often vague, so that it remains a human task to execute them. Automating idea generation and implementation in one coherent system would significantly shift the role of humans in the scientific process. Here we present \AIMandel, an LLM agent that can generate and implement ideas in quantum physics. \AIMandel formulates ideas from the literature and uses a domain-specific AI tool to turn them into concrete experiment designs that can readily be implemented in laboratories. The generated ideas by \AIMandel are often scientifically interesting -- for two of them we have already written independent scientific follow-up papers. The ideas include new variations of quantum teleportation, primitives of quantum networks in indefinite causal orders, and new concepts of geometric phases based on closed loops of quantum information transfer. \AIMandel is a prototypical demonstration of an AI physicist that can generate and implement concrete, actionable ideas. Building such a system is not only useful to accelerate science, but it also reveals concrete open challenges on the path to human-level artificial scientists.
\end{abstract}

\section{Introduction}
Artificial intelligence (AI) techniques are becoming increasingly important in scientific research \cite{krenn2022scientific, wang2023scientific,musslick2025automating} as tools to accelerate or even enable different research activities. In physics, these tasks range from the design of new experiments \cite{krenn2016automated, krenn2021conceptual, krenn2020computer}, to automated discovery of governing equations \cite{udrescu2020ai,cranmer2023interpretable}, symmetry \cite{wetzel2020discovering,yang2024symmetry} and phase \cite{carrasquilla2017machine} from data, to speed up \cite{dax2025real} and scientific control \cite{degrave2022magnetic,buchli2025improving} of large scientific instruments and for potential foundational models of physics\cite{ohana2024well,barman2025large}. With the emergence of large language models (LLMs) and LLM agents, more parts of the scientific workflow have been automated. An example in chemistry is the planning and execution of synthesis routes for new materials \cite{boiko2023autonomous,m2024augmenting,zou2025agente} and uncovering new materials \cite{wang2024efficient,ghafarollahi2025sciagents} (for a more comprehensive review on applications of LLM agents in chemistry see \cite{ramos2025review}); in physics, the rediscovery of underlying equations from data \cite{nagele2025agentic}, the control in quantum labs \cite{cao2025automating} or the design of specific quantum or photonic circuits~\cite{yu2025quasar,sharma2025ai} or metamaterials \cite{lu2025agentic};  in mathematics, by using powerful coding agents \cite{novikov2025alphaevolve}, AI can contribute to new mathematics \cite{georgiev2025mathematical}.

All of these techniques have in common that the original research question and task have been defined by the human operators prior to the execution of the AI-driven tools. A critical question is whether machines could automatically discover new scientific directions or research ideas. With access to large amounts of scientific papers, scientists were able addresse this question. One way is to create evolving knowledge graphs of science -- for example biochemistry \cite{rzhetsky2015choosing} to quantum physics \cite{krenn2020predicting}, AI \cite{krenn2023forecasting} and chemistry \cite{marwitz2025predicting} --, which allows to model the past progress of research ideas and predict future research directions and even the impact of future research \cite{shi2023surprising,gu2025forecasting}. With the emergence of large-language models, it became possible to formulate research ideas directly in natural language, and evaluate these ideas and hypotheses by human experts \cite{wang2024scimon, gu2024interesting, yang2024large, si2024can, baek2025researchagent}. One drawback of these studies is that, while the AI-generated ideas might be interesting, it is often unclear how to actually implement and use them.

Therefore, one crucial question is how to develop automated systems that can generate new creative scientific ideas and automatically execute them. Such a system would automate a large fraction of the scientific process, and its results might be particularly novel and surprising for domain experts because no human was involved in the idea generation and execution, and they are only presented with the final results. This goal is sometimes framed \textit{AI scientist} or \textit{artificial scientist} \cite{xie2025far, lu2025can}. Recently, numerous prototype \textit{AI scientists} have emerged, for example in biochemistry \cite{gottweis2025towards, mitchener2025kosmos}, computer science \cite{lu2024ai} or general science \cite{villaescusa2025denario}. These systems have long, complex reasoning traces and can produce useful scientific results -- yet they are still following a concrete scientific question defined by humans. 

Here, we present \AIMandel \footnote{\AIMandel{} is named in tribute to Leonard Mandel, an early pioneer of quantum optics whose ideas still keep inspiring physicists.}, an LLM agent system\footnote{The LLM used in the final version is \textit{OpenAI's} reasoning model o4-mini.} that finds and demonstrates new, interesting ideas in quantum physics by accessing scientific literature and intelligent scientific discovery tools. \AIMandel does not have a concrete human-defined task beyond finding interesting ideas in quantum physics. From there, it develops ideas using scientific papers and formulates execution strategies for quantum optics lab by accessing \pytheus \cite{ruiz2023digital} -- an AI-driven tool for the design of new quantum experiments. \AIMandel has to reformulate its conceptual ideas in natural language into a computer language that can be understood by \pytheus. \AIMandel can produce high-quality, publishable contributions to quantum physics, which we demonstrate by writing two scientific papers based on the generalized results of \AIMandel \cite{arlt2025automated,arlt2025automated2}. The ideas range from new constructions of non-local quantum states to quantum transformations with different resources than those that have been used so far, to the construction of new phases via information transfer. \AIMandel is a prototype of an \textit{AI physicist} framework with access to scientific literature and intelligent generative tools that can produce high-quality research level ideas and results.

We will first explain a big-picture idea of \AIMandel, before we describe the intricate construction of the LLM agents. Afterwards we will explain seven concrete ideas from \AIMandel and show how we used two of them to write independent scientific papers. Finally, we will discuss what is missing to reach true artificial scientists, and how we might be able to get there.

\begin{figure}
    \centering
    \includegraphics[width=0.49\textwidth]{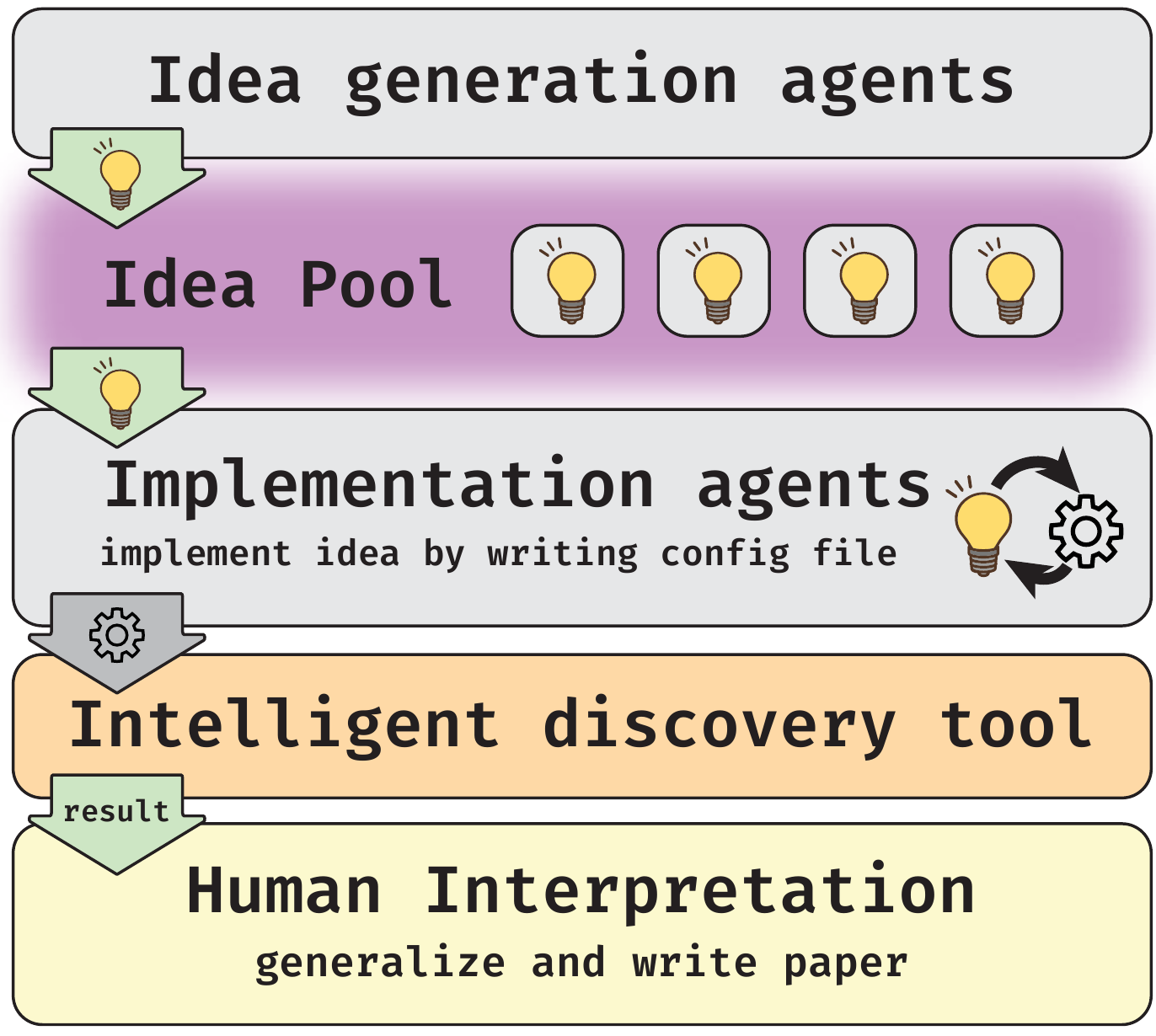}
    \caption{\textbf{Workflow of \AIMandel.} Agents propose ideas, query literature for overlap, and interface with \pytheus{} for implementation. Successful designs are evaluated by human experts and the top results are developed into a research project and published.}
    \label{fig:workflow}
\end{figure}

\begin{figure*}[ht!]
    \centering
    \includegraphics[width=0.95\textwidth]{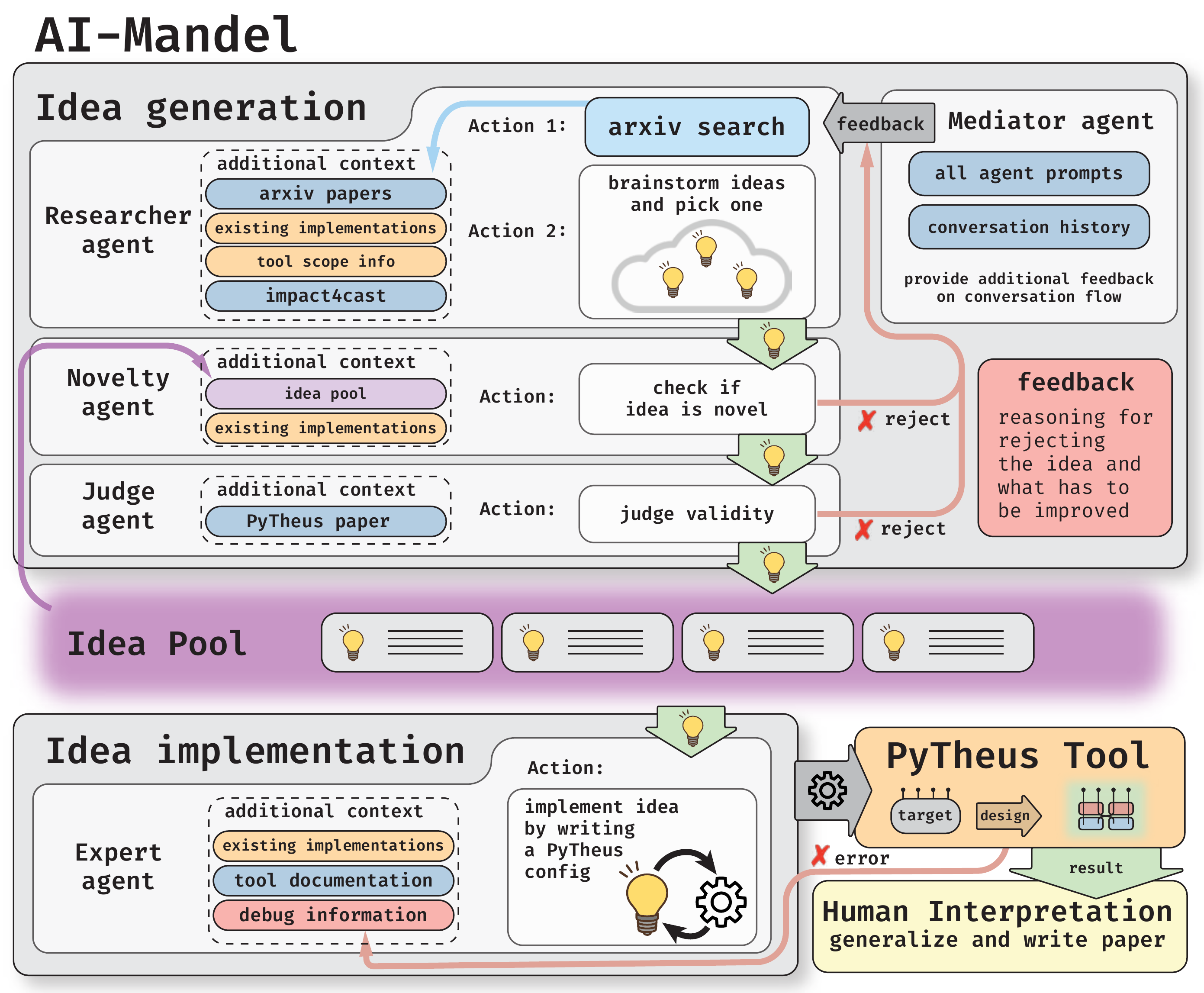}
    \caption{\textbf{Detailed Overview of Agent Interactions.} The system is split into \textit{Idea generation} and \textit{Idea implementation}. \textit{Idea-generation} consists of four agents (\textit{Researcher}, \textit{Novelty}, \textit{Judge}, and \textit{Mediator}).  The \textit{Researcher} is prompted to come up with an interesting target to search for through a design tool for quantum optics experiments.  The prompt of the \textit{Researcher} contains (i) the abstracts of three random papers from the arXiv dataset of quantum physics articles, (ii) documentation and implementations of existing \PyTheus design queries., (iii) additional information about limitations on type of experiments that can be searched for, (iv) a pair of quantum physics concepts that currently are not combined but predicted to have potential to be impactful by \impactfourcast with the request to combine them in the final idea. The \textit{Researcher} gets to choose between two actions. Action 1: Formulate an arxiv query to search literature. The first three abstracts matching the query are added to the context. Action 2: Formulate three possible ideas and pick one of them, reasoning about novelty and feasibility. The idea is passed on to the \textit{Novelty agent}, which is prompted to \textit{accept} or \textit{reject} suggestions based on novelty with respect to existing ideas generated by the agents (\textit{Idea Pool}) or existing examples which have been implemented in \pytheus and published before. If the \textit{Novelty agent} rejects the suggestion, it is tasked to give feedback to the \textit{Researcher}, which will attempt to improve its suggestion until the \textit{Novelty agent} accepts or a maximum number of iterations is reached. In the case of the \textit{Novelty agent} accepting the suggestion, it is passed on to the \textit{Judge}, which is tasked to \textit{accept} or \textit{reject} suggestions based on feasibility. In the case of rejection, this agent also gives feedback on flaws in the current suggestion to the \textit{Researcher}. When an idea is accepted by both \textit{Novelty agent} and \textit{Judge},  we consider it to be successfully generated and it is stored in the \textit{Idea Pool}. The \textit{Mediator agent} is called every third iteration and is tasked to highlight inefficiencies in the conversation between the other three agents. Its prompt contains the prompts of all other agents and their conversation up to this point. For \textit{Idea implementation} the \textit{Expert} is prompted to write a working config file for the \textit{PyTheus Tool}. Error messages are passed back to the \textit{Expert} for debugging with the task to fix the implementation. Successful designs are stored and passed on to human domain experts.}
    \label{fig:agent-tool}
\end{figure*}
\begin{figure*}[ht!]
    \centering
\includegraphics[width=1\textwidth]{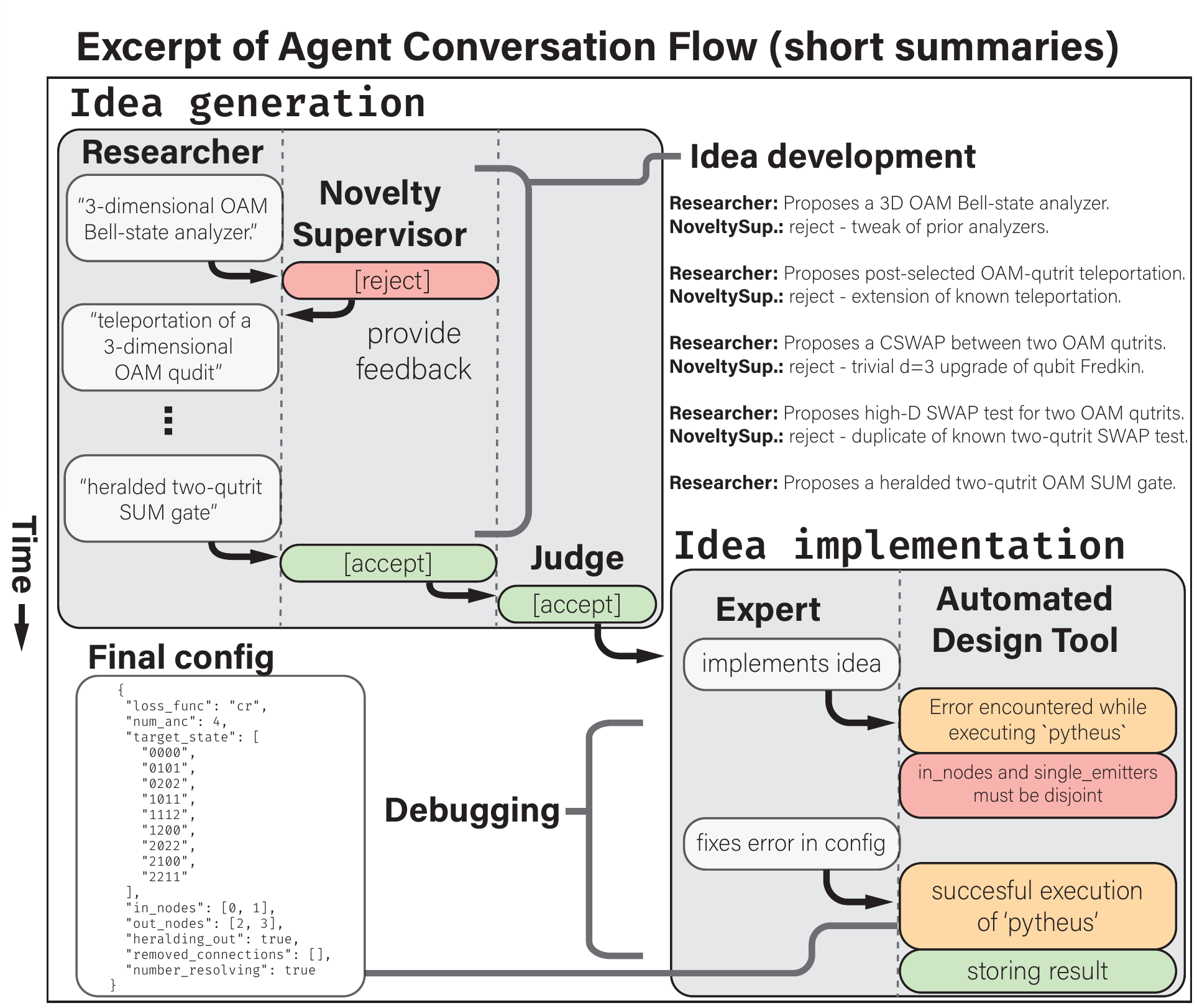}
    \caption{\textbf{Example Agent Interaction.} The initial \textit{Researcher} proposal is rejected by the \textit{Novelty Supervisor}. After multiple iterations, the modified proposal is accepted by the \textit{Novelty Supervisor} and the \textit{Judge}. The initial attempt by the \textit{Expert} does not succeed. Upon fixing the parameter settings, the \textit{Expert} executes the tool successfully, and the finalized design is subsequently stored.}
    \label{fig:example}
\end{figure*}

\section{\AIMandel workflow}
\textbf{The big-picture} -- The big-picture goal of \AIMandel is to discover novel ideas in quantum optics and concrete blueprints for their practical implementation. \AIMandel consists of two LLM agent systems\footnote{The code and prompts used in our implementation of \AIMandel can be found in our \href{https://github.com/artificial-scientist-lab/ai-mandel}{github repository}.} (each consisting of sub-agents; each LLM is an independent instance of \textit{OpenAI's} reasoning model o4-mini \cite{openai_o3_o4mini_systemcard_2025}) with access to an intelligent discovery tool \pytheus, see Fig.~\ref{fig:workflow}. \pytheus can discover concrete experimental implementations (often using new unexplored methods) for concrete research questions provided in \cite{ruiz2023digital}. \AIMandel's task is to invent new research ideas (using the \textit{idea generation agents}) -- fully accepted ideas are stored in the \textit{idea pool}. The \textit{implementation agents} take ideas from the idea pool (written in natural language) and translate them into instructions for \pytheus. \pytheus executes the instructions -- if unsuccessfully, it returns errors back to the implementation agent; if successful, it reports the natural language descripion with the code implementation and the final solution to the human scientists. We went through 184 successfully implemented ideas and confirmed they contain many interesting results for experimental quantum optics. We show seven of these ideas later in the text and have used two of them to develop scientific papers \cite{arlt2025automated, arlt2025automated2}.
\begin{figure*}[!t]
    \centering
\includegraphics[width=1\textwidth]{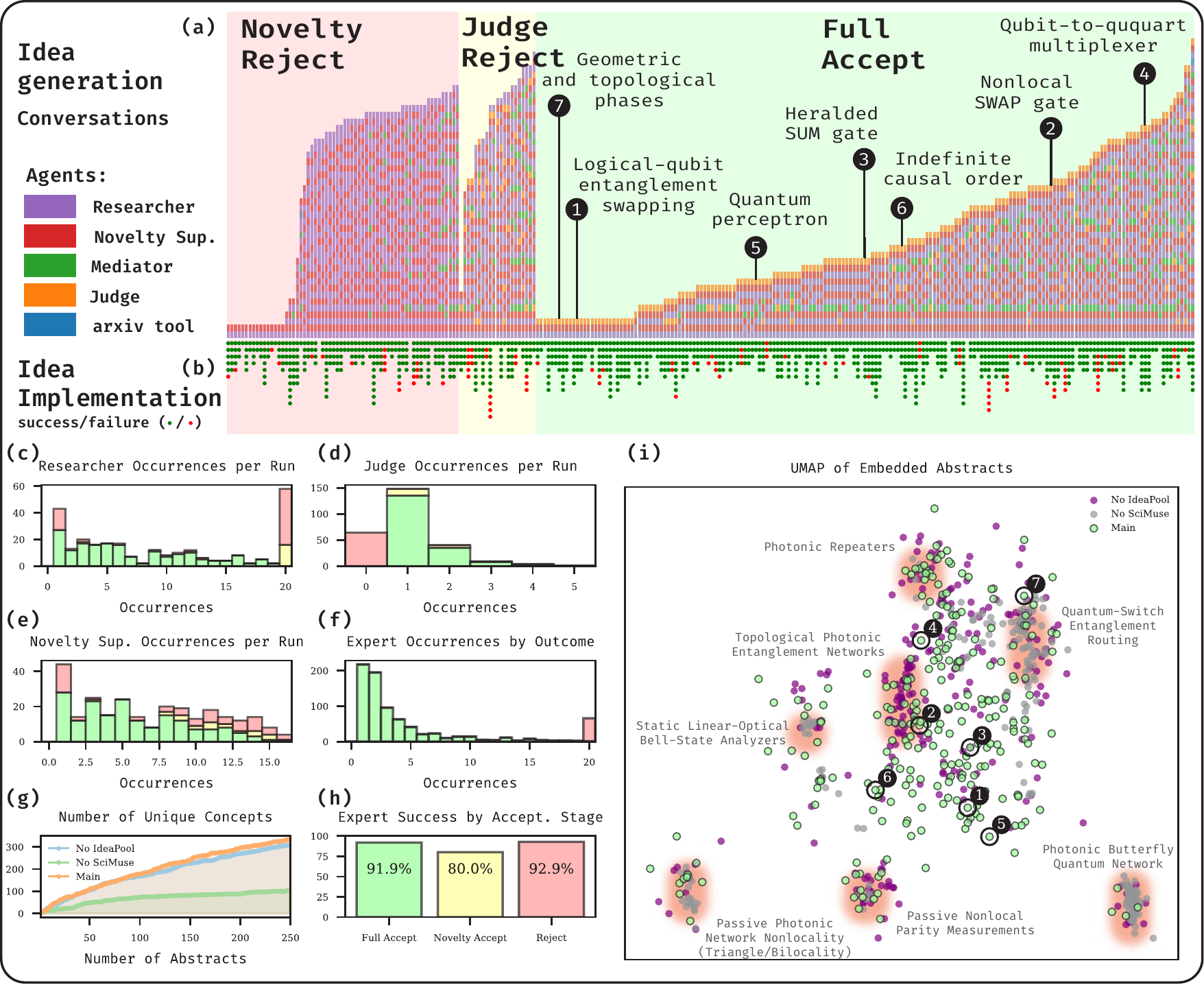}
    \caption{\textbf{Analysis of Agent Interaction --} (a) Visualization of all idea generation conversations. 'Full Reject' are ideas that did not get past the Novelty Supervisor during multiple iterations. 'Novelty Accept' are ideas that were accepted by the Novelty supervisor, but did not pass the Judge during multiple iterations. 'Full Accept' are ideas that passed both filters. We label the 7 final ideas by numbers.  (b) Showing successful (green) and unsuccessful (red) implementation attempts for each idea. Each rectangle segment is one run. To test the implementability of 'Full Reject' and 'Novelty Accept', the Researcher suggestions were passed to the Expert despite the rejections. (c)-(e) Histograms showing Occurrences of the idea generation agents (Researcher, Judge, Novelty Supervisor). x-axis: how often an agent appears, y-axis: number of runs for each bin. The color identifies 'Full Reject', 'Novelty Accept', or 'Full Accept'. (f) Histogram showing how often the Expert is called during an implementation run. Red: unsuccessful implementations. (g) Cumulative sum of new \scimuse concepts mentioned in the abstracts written by the Researcher. 'Main' is our main run of 'Full Accept' ideas with \scimuse and Idea Pool. (h) Expert success rate (reaching a successful PyTheus run) for different levels of success in idea-generation (i) 748 abstracts (from Main, No \scimuse, and No Idea Pool) are embedded using \texttt{text-embedding-3-large} (text embedding model by OpenAI) and approximated in a two dimensional space using UMAP and PCA. We label the 7 final ideas by numbers. We manually identified some clusters (especially prevalent in No Idea Pool and No \scimuse) and analyzed the contained abstracts, giving them a comprehensive title.
    }
    \label{fig:data}
\end{figure*}

\textbf{Workflow details} -- 
The details of the workflow are described in Fig.\ref{fig:agent-tool} and its caption. The idea generation system consists of three inter-connected agents. The first (researcher agent) gets input about \pytheus, and for diversity, three random abstracts of quantum physics papers on arXiv, as well as two concepts from \texttt{SciMuse}'s \impactfourcast \cite{gu2024interesting,gu2025forecasting}, a machine-learning system with access to millions of scientific papers and trained to predict highly impactful yet unexplored research directions. The researcher agent also has direct access to arXiv to verify the novelty of ideas. The researcher agent then generates ideas that are then judged by two judge-agents, that evaluate the novelty (with access to the previously generated ideas and access to existing implementation from the literature) and implementability via \pytheus (via access to the full \pytheus paper and extra relevant examples). Both judge-agents can give feedback to the researcher agent, who adapts its idea accordingly. A mediator agent is used to prevent unsuccessful communication attempts (e.g. when the agents do not change their output despite feedback or new input). If the idea was successfully generated, it is added to the idea pool. In our experiments, we generated in this way 187 ideas in the pool.

The implementation agent then takes a natural-language idea from the pool (randomly selected), and translates it into concrete instructions that can be read by \pytheus. \pytheus can return error messages to the implementation agent, who will adapt the instruction files accordingly. If \pytheus succeeds, the final solutions -- including the natural langauge idea, the instruction file for \pytheus and the resulting experimental blueprint are stored for the human scientists to investigate.

A typical run of \AIMandel can be seen in Fig.\ref{fig:example}. There, the novelty supervisor agent rejects an initial idea of the researcher agent. The researcher modifies its suggestion and the idea is subsequently accepted by both the novelty supervisor and the judge . Then, the expert agent attempts to write instructions for \pytheus, which fails in the first attempt but -- with feedback about the error -- succeeds eventually. The final result of this run is one of the seven interesting examples we explain below.

In Fig.\ref{fig:data}, we show details of the interactions between the agents and their tools. We show the statistics of the 187 generated ideas by the idea-generation agents, and the 804 implementations (739 were successful while 65 were unsuccessful; 184 of the 187 ideas have been successfully implemented at least once). We also demonstrate the diversity of the generated ideas and confirm with two different metrics that both \impactfourcast concepts and access to the previously generated ideas in the idea pool contributed to the diversity of the results.

\section{Results}
\begin{figure*}[!t]
    \centering
    \includegraphics[width=0.99\textwidth]{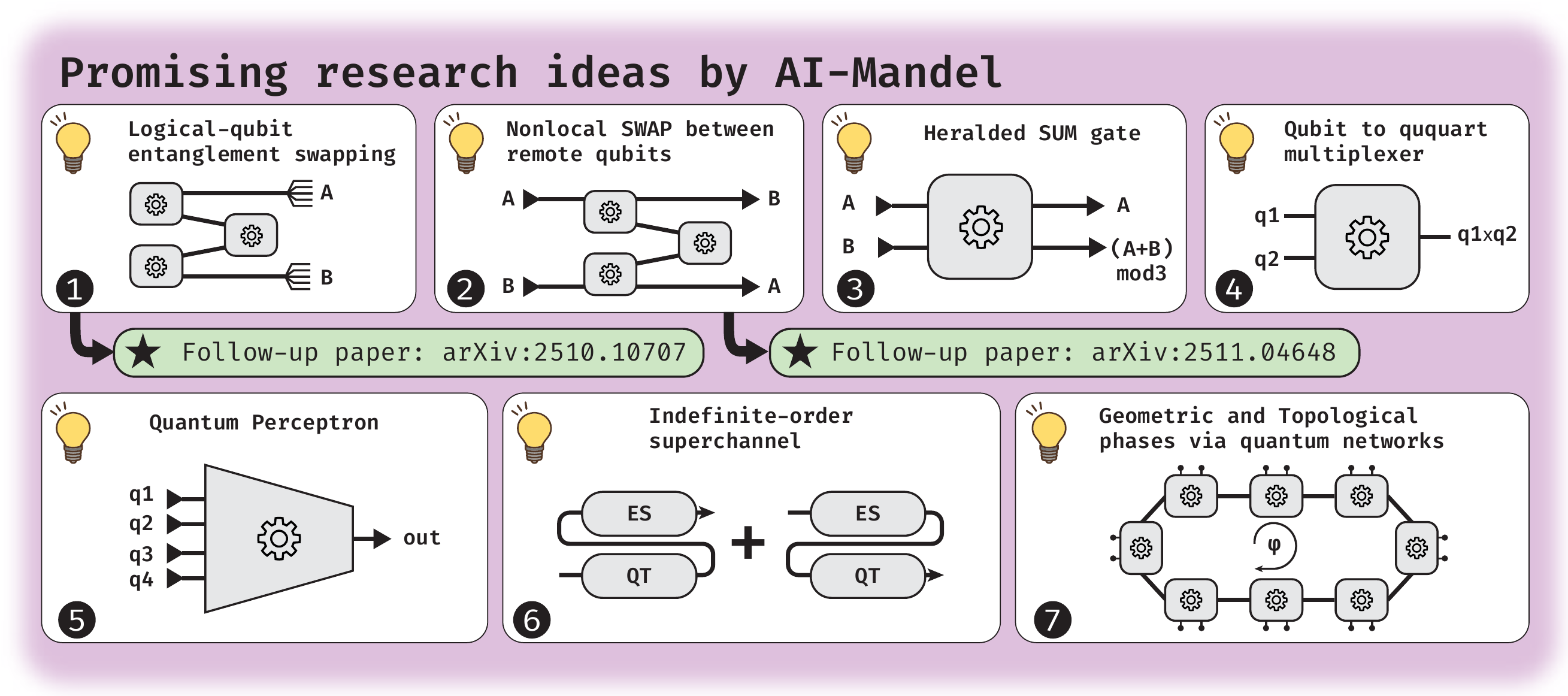}
    \caption{\textbf{Human filtered ideas.} We show the seven implementations generated by the agents that we considered most viable to be developed into a full research project. We give detailed descriptions of these ideas in the results section.}
    \label{fig:results}
\end{figure*}

In this section we give a short description for seven ideas generated by \AIMandel that we considered most interesting. Each of them is also illustrated in Fig.~\ref{fig:results}. For two of them (Result 1 and 2), we have written independent scientific papers demonstrating that the core ideas discovered by \AIMandel are of high quality\footnote{We believe that many other of \AIMandel's ideas could be expanded into viable research papers or thesis projects for students.}. 

\noindent
\textbf{Result 1 -- Logical-qubit entanglement swapping:}
Result 1 discusses the idea of quantum entanglement swapping \cite{pan1998experimental} of logical entanglement. Recently, it was discovered and experimentally demonstrated that a new form of quantum operation, that mimics core properties of entanglement swapping, can be implemented without shared entanglement and without Bell state measurements -- the core resources believed necessary for entanglement swapping \cite{wang2024entangling}. Instead, a technique directly inspired by the works of Leonard Mandel \cite{wang1991induced} can be used as a resource. \AIMandel decided to expand this technique into the swapping of logical entanglement -- a core property of quantum states in error-corrected quantum computers. This proposal is, to the best of our knowledge, new and immediately actionable and generalizable. Therefore, we decided to generalize the core ideas and describe them in detail in an independent scientific paper \cite{arlt2025automated}. Specifically, in the article we first demonstrate the idea in simpler systems until we show -- as a final example, the full idea and implementation created by \AIMandel.

\noindent
\textbf{Result 2 -- Non-local SWAP between remote qubits:}
A SWAP gate exchanges the states of two qubits, but typically requires both to be in the same location. Quantum information theory shows that such operations can also be realized non-locally, by consuming shared entanglement and performing joint measurements at an intermediate station \cite{yi2002implementation}. \AIMandel proposes a photonic implementation of a non-local SWAP between two remote qubits, held by Alice and Bob, without sharing entanglement between the parties, but instead using the indistinguishability of photon pair origins \cite{hochrainer2022quantum}. When analyzing the solution, we understood that the idea is immediately generalizable, and -- by accident, discovered that a core component of \AIMandel's solution is a new form of quantum teleportation that uses fundamentally different resources compared to the standard protocol \cite{bennett1993teleporting, bouwmeester1997experimental}. We describe all of these findings based on \AIMandel's ideas in another, independent scientific paper \cite{arlt2025automated2}.

\noindent
\textbf{Result 3 -- Heralded SUM gate:}
Entangling operations, particularly controlled-NOT (CNOT) gates, are essential components of quantum computation, enabling universal logic when combined with single-qubit operations. Over the past two decades, a series of photonic implementations has demonstrated the feasibility of \emph{heralded} entangling gates, in which ancillary detections certify successful operation while leaving the output photons available for further processing~\cite{gasparoni2004realization, huang2004experimental, zhao2005experimental, bao2007optical, li2021heralded}. Extending such operations to \emph{high-dimensional} systems is especially appealing, as qudit encodings enlarge the accessible Hilbert space and reduce the number of required gates. The high-dimensional generalization of the CNOT is the \emph{controlled-SUM (CSUM)} gate, which performs modular addition on the target conditioned on the control state. The design of such gates has also been explored through digital approaches~\cite{ruiz2023digital, gao2020computer} and experimentally implemented ~\cite{imany2019high,meng2024experimental}, which employ \emph{postselected} operations using single-photon or two-dimensional control. \AIMandel proposes a photonic implementation of a heralded CSUM gate between two photonic \emph{qutrits}. In contrast to previous implementations, the gate operates in a heralded fashion on two distinct multi-level photons. It employs interference between indistinguishable pair-creation origins and does not require Bell state measurements. Conceptually, this mechanism can be generalized to arbitrary dimension, providing a pathway toward heralded entangling operations for photonic qudits. This idea inspired part of the content of our scientific paper \cite{arlt2025automated2}.

\noindent
\textbf{Result 4 -- Qubit to ququart multiplexer:}
Quantum information can be encoded either across multiple particles or within multiple degrees of freedom of a single particle. Converting between these encodings enables more flexible use of quantum channels and resources \cite{passaro2013joining,vitelli2013joining}. \AIMandel proposes a multiplexer that takes in a two-qubit state and outputs an equivalent ququart state, effectively switching between particle-based and mode-based encodings. The required resources are fundamentally different than in the existing literature, which could lead to alternative strategies in quantum network architectures.

\noindent
\textbf{Result 5 -- Quantum Perceptron:}
The perceptron, the basic unit of classical neural networks, combines multiple inputs through a weighted sum and applies a non-linear activation function. Implementing an analogue in quantum systems is difficult because unitary dynamics are intrinsically linear. Recent proposals have suggested that measurement and post-selection can provide the required non-linearity in quantum settings \cite{torrontegui2019unitary,tacchino2019artificial}.
\AIMandel proposes a measurement-induced quantum perceptron realized in linear optics which has not been discussed in the literature before. Two photonic qubit inputs are combined with entangled ancillas, and post-selection heralds a single output qubit. This approach could be extended to multi-qubit inputs or to higher-dimensional (qudit) perceptron gates, offering a route toward photonic realizations of quantum neural networks.  

\noindent
\textbf{Result 6 -- Indefinite-order superchannel:}
Quantum mechanics permits not only the superposition of different events, but even the superposition of the order in which two events take place, a phenomenon known as indefinite causal order \cite{chiribella2013quantum, oreshkov2012quantum}. This has been demonstrated in photonics for simple gates, where the order of applying the same operation is put into superposition \cite{procopio2015experimental,rubino2017experimental, rozema2024experimental}. \AIMandel extends this concept to full quantum network protocols by proposing a photonic experiment that coherently combines entanglement swapping and quantum teleportation. An ancilla photon prepared in a superposition controls whether the input qubit undergoes \textit{swap then teleport} or \textit{teleport then swap}. With the ancilla in a balanced state, both orders occur simultaneously, yielding a coherent superposition of outcomes. The fundamental building blocks of quantum networks in a superposition of orders could lead to new insights into problems of indefinite causal order.

\noindent
\textbf{Result 7 -- Geometric and Topological phases via quantum teleportation networks:}
Nontrivial geometric or topological phases arise when a quantum system is transported around a closed loop in an appropriate space, e.g. physical space, momentum space or parameter space. A famous example is the Berry phase \cite{berry1984quantal,xiao2010berry}: if a spin-½ is moved slowly by a magnetic field whose direction traces a loop on the Bloch sphere, the spin returns with a phase set by half the solid angle enclosed which is visible as a shift in interference patterns. Another key topological effect is the Aharonov–Bohm effect, a topological effect that occurs where an electron encircling an isolated magnetic flux acquires a phase that depends only on the enclosed flux, not on the detailed path. The same Berry-phase geometry, accumulated over loops in momentum space, underpins topological phases of quantum matter and their applications \cite{hasan2010colloquium}, including topological qubits.

\AIMandel now proposes to transport a quantum state around a closed loop not in physical space, but by teleporting \cite{bennett1993teleporting,bouwmeester1997experimental} or entanglement-swapping \cite{zukowski1993event} it sequentially between nodes arranged in a ring (e.g., $A \to B \to C \to A$) \cite{renou2019genuine,baumer2025exploring}. Depending on how these link resources (entangled pairs, measurement bases, or protected code spaces) are engineered, the loop action could realize a smoothly tunable geometric phase or a quantized, perturbation-robust topological phase. The precise resource design that guarantees a nontrivial loop phase remains an exciting research question for the future. To the best of our knowledge, the idea of using standard quantum-information protocols to engineer geometric or topological phases by closing a loop in network space has not been described before.

%\begin{figure}
%    \centering
%    \includegraphics[width=0.49\textwidth]{figs/fig1.pdf}
%    \caption{\textbf{Positioning of \AIMandel in the AI for physics pipeline.} The pipeline comprises four stages: \emph{idea generation}, \emph{idea implementation}, \emph{generalization of results} (clustering related designs to extract common mechanisms), and \emph{interpreting \& writing} (placing findings in a coherent storyline and situating them within prior work). \AIMandel primarily automates the first two stages: it proposes candidate quantum-optics experiments and translates them into executable \pytheus{} configurations, iterating on tool feedback. The validated designs—and informative failures—then feed downstream synthesis and, ultimately, human-led interpretation and manuscript writing.}
    %\label{fig:overall}
%\end{figure}

\section{Conclusions}
\AIMandel is an early instance of an AI physicist that can generate new ideas in quantum physics and implement these ideas using external intelligent tools. At the core, it is a system of LLM agents (based on \textit{OpenAI’s} models) with access to \pytheus, an AI framework to discover new quantum experiments. \AIMandel and its future successors are prototype AI scientists that help us to understand what can readily be automated, and which parts of the automation of science remains a challenge\footnote{\textit{What I cannot create, I do not understand}, Richard Feynman.}.

\textbf{Extending the set of intelligent tools} -- At the moment, \AIMandel has access to one concrete AI-based tool in quantum optics -- \pytheus \cite{ruiz2023digital}. Many other tools could be added to its toolbox, which would add general capabilities and enable future versions of \AIMandel to generate and execute creative ideas from a larger pool of possibilities. Examples include \texttt{PennyLane} with the ability to simulate and optimize photonic circuits \cite{bergholm2018pennylane}, \texttt{Qiskit} for the optimization of quantum computing circuits \cite{qiskit2024}, \texttt{Quanundrum} for simulating thought experiments with quantum agents \cite{nurgalieva2022thought}.

\textbf{Automated Generalizations} -- Future versions of \AIMandel could have the ability to autonomously generalize the discovered results. This ability could be implemented by applying the concept of \textit{meta-design} \cite{arlt2024meta}, where the agent does not produce a single solution, but produces computer code that solves a general class of problems.

\textbf{Automated Interpretation and Conceptualization} -- At this stage, even though a future AI physicist is able to find and implement ideas in diverse fields of (quantum) physics, and is able to generalize them autonomously, it still lacks abilities that human physicists have. For example, humans interpret results against the background of an existing corpus of knowledge and understand the underlying principles \cite{krenn2022scientific,de2017understanding,barman2024towards}. For example, when we analyzed one result by \AIMandel in detail, we understood that in order to solve its question, it invented a new variant of quantum teleportation that has never been discussed before \cite{arlt2025automated2}. \AIMandel did not have the ability to understand this connection yet, and it will be exceptionally interesting to know how it could find these insights autonomously.

\textbf{Fully automated scientists} -- While we believe \AIMandel is a crucial step towards the automation of science, many questions remain open. For example, the way \AIMandel estimates novelty and attempts to increase the diversity of ideas is different from human scientists \cite{champion2025strong}. Humans follow internal motivations such as interest \cite{silvia2008interest}, curiosity \cite{de2024wonderstruck}, and the desire of surprise \cite{ivanova2024surprise}. It might be valuable or even necessary to understand these genuine traits of human scientists so we can artificially recreate them and build true artificial scientists.

\section*{Acknowledgments}
The authors thank Shanmugapriya Kanagasabapathi and Jonathan Klimesch for their useful comments on the manuscript.
MK acknowledges support by the European Research Council (ERC) under the European Union’s Horizon Europe research and innovation programme (ERC-2024-STG, 101165179, ArtDisQ) and from the German Research Foundation DFG (EXC 2064/1, Project 390727645). X.G. acknowledges support from the NOA Collaborative Research Center and the Alexander von Humboldt Foundation.  

% References
%\bibliography{main}
%\bibliographystyle{unsrtnat}

\printbibliography

\onecolumn
\newpage
\appendix
\section{Ideapool Examples (Final selection)}
We show the abstracts written by the \textit{Researcher}, summarizing its ideas before adding them to the \textit{Idea Pool}. More generated abstracts and the logs for the corresponding runs can be found in our \href{https://github.com/artificial-scientist-lab/ai-mandel}{github repository}.
\begin{myverbatim}
Title:  
Heralded Linear‐Optical Two-Qutrit SUM Gate (mod 3) in Laguerre–Gaussian OAM Modes  

Mini-abstract:  
We present a Pythous-optimized, heralded linear-optical network that implements the two-qutrit SUM gate U_S∶|i⟩_A|j⟩_B → |i⟩_A|i⊕j (mod 3)⟩_B on single photons encoded in Laguerre–Gaussian OAM modes ℓ∈{−1,0,1}.  Using only static beam splitters, phase shifters, two SPDC ancilla sources and number-resolving detectors—with a unique four-click ancilla pattern to herald success—this is the first explicit high-dimensional entangling gate realized in the OAM platform.
\end{myverbatim}

\begin{myverbatim}
Title: Entanglement Swapping of 2×2 Toric-Code Logical Qubits via Linear Optics

Abstract: We demonstrate entanglement swapping between two distant logical qubits, each encoded in a minimal 2×2 toric-code patch of four photons at nodes A and D, by performing a post-selected logical Bell measurement on intermediate patches B and C using only probabilistic photon-pair sources, passive linear optics, and ancilla detection. The successful eight-fold coincidence projects detectors 0–3 and 12–15 onto the unnormalized target quantum state  
["00000000", "00001111", "11110000", "11111111", "00110011", "00111100", "11000011", "11001100"],  
heralding a fault-tolerant logical Bell pair between A and D without any feed-forward correction.
\end{myverbatim}

\begin{myverbatim}
Title: Heralded Non-local SWAP Gate Between Remote Photonic Qubits

Mini-Abstract: We target the realization of a fully heralded non-local SWAP gate between two unknown photonic qubits |ψ⟩ₐ at node A and |φ⟩_B at node B using only static linear optics, probabilistic Bell-pair ancilla sources, and post-selected detection. A central four-photon interferometric measurement on ancilla photons heralds the successful one-shot exchange of the unknown qubits between output modes A_out and B_out, without any dynamic feed-forward.
\end{myverbatim}

\begin{myverbatim}
Title: Telecom-Band Frequency-Bin Teleportation MUX for Qubit→Ququart Isometry

Mini-Abstract: We propose a fully passive, measurement-assisted linear-optical circuit that fuses two qubits, each encoded in a Kerr microcomb’s telecom-band frequency bins, into a single ququart channel via teleportation.  Using two probabilistic Bell-pair photon sources, ancilla modes, a static interferometer and photon-number-resolving detectors, the heralded Choistate  
∑_{i,j∈{0,1}} |i,j⟩_in |2i+j⟩_out  
realizes the isometry mapping |00⟩→|0⟩, |01⟩→|1⟩, |10⟩→|2⟩ and |11⟩→|3⟩ without dynamic feed-forward or cavities.
\end{myverbatim}

\begin{myverbatim}
Title: Heralded Linear-Optical Quantum Perceptron Activation Gate

Mini-Abstract: We implement a measurement-induced non-linear "quantum perceptron"gate in a fully passive linear-optical network by combining two unknown photonic qubit inputs with Bell-pair ancillas and post-selection heralding. The target is a single output qubit at node 2 whose probability of measuring |1⟩ realizes a tunable threshold (or sigmoid) function f(w₁·x₁ + w₂·x₂), providing a genuine non-unitary activation primitive for photonic quantum machine learning.
\end{myverbatim}

\begin{myverbatim}
Title: Continuous Topological Phase Transition of a Kitaev 1D Majorana Chain via Bell-Measurement Tuning in a Linear-Optical Entanglement-Swapping Network

Mini-abstract: We propose and simulate in Pytheus a linear-optical quantum network of probabilistic photon-pair sources and intermediate Bell-state measurements whose beam-splitter reflectivity r acts as a continuous control parameter, reproducing the continuous topological phase transition of the target Kitaev 1D Majorana chain. As r crosses a critical threshold r_c, long-range end-to-end entanglement between edge nodes A and Z suddenly emerges, faithfully mapping onto the onset of Majorana zero-mode correlations in the original topological model.
\end{myverbatim}

\begin{myverbatim}
Title: On-Chip Indefinite-Order Photonic Superchannel Switch

Abstract: We target the design of an integrated-photonic "superchannel switch"that coherently superposes two distinct quantum network primitives—entanglement swapping (ES) and quantum teleportation (QT)—in indefinite causal order. By combining two probabilistic Bell-pair sources, a single-photon input qubit and a control ancilla in a fixed linear-optical interferometer, we perform Bell-state measurements and post-select on "+"outcomes to herald a coherent superposition of the ES→QT and QT→ES sequences, all without active feed-forward.
\end{myverbatim}

\newpage
\section{Implementation results}
We show the final config files produced by \AIMandel for the seven results described in the main text. All config files produced by the Expert agent and the corresponding logs can be found in our \href{https://github.com/artificial-scientist-lab/ai-mandel}{github repository}.
\begin{myverbatim}
{
  "description": "Heralded SUM gate (mod 3) on two OAM qutrits",
  "foldername": "sum_qutrit_mod3",
  "bulk_thr": 0.1,
  "edges_tried": 30,
  "ftol": 1e-06,
  "loss_func": "cr",
  "num_anc": 4,
  "num_pre": 1,
  "optimizer": "L-BFGS-B",
  "imaginary": false,
  "safe_hist": true,
  "samples": 10,
  "target_quantum": [
    "0000",
    "0101",
    "0202",
    "1011",
    "1112",
    "1200",
    "2022",
    "2100",
    "2211"
  ],
  "in_nodes": [0, 1],
  "out_nodes": [2, 3],
  "thresholds": [0.3, 0.1],
  "heralding_out": true,
  "single_emitters": [],
  "amplitudes": [],
  "tries_per_edge": 5,
  "removed_connections": [],
  "number_resolving": true
}
\end{myverbatim}
\begin{myverbatim}
{
  "description": "Logical-qubit entanglement swapping in minimal 2x2 toric code",
  "foldername": "ES_toriccode",
  "bulk_thr": 0.01,
  "edges_tried": 30,
  "ftol": 1e-05,
  "loss_func": "cr",
  "num_anc": 8,
  "num_pre": 1,
  "optimizer": "L-BFGS-B",
  "imaginary": false,
  "safe_hist": true,
  "samples": 10,
  "target_quantum": [
    "00000000", "00001111", "11110000", "11111111",
    "00110011", "00111100", "11000011", "11001100"
  ],
  "in_nodes": [],
  "out_nodes": [],
  "thresholds": [0.3, 0.1],
  "heralding_out": null,
  "single_emitters": [],
  "amplitudes": [],
  "tries_per_edge": 5,
  "removed_connections": [
    [0,8],[0,9],[0,10],[0,11],[1,8],[1,9],[1,10],[1,11],
    [2,8],[2,9],[2,10],[2,11],[3,8],[3,9],[3,10],[3,11],
    [0,12],[0,13],[0,14],[0,15],[1,12],[1,13],[1,14],[1,15],
    [2,12],[2,13],[2,14],[2,15],[3,12],[3,13],[3,14],[3,15],
    [4,12],[4,13],[4,14],[4,15],[5,12],[5,13],[5,14],[5,15],
    [6,12],[6,13],[6,14],[6,15],[7,12],[7,13],[7,14],[7,15]
  ],
  "seed": null,
  "unicolor": false,
  "number_resolving": true,
  "novac": null,
  "loops": null,
  "topopt": null,
  "dimensions": [],
  "brutal_covers": null,
  "verts": [],
  "anc_detectors": []
}
\end{myverbatim}

\begin{myverbatim}
{
  "description": "Heralded nonlocal SWAP between remote qubits A and B",
  "foldername": "remote_swap",
  "bulk_thr": 0.1,
  "edges_tried": 30,
  "ftol": 1e-06,
  "loss_func": "cr",
  "num_anc": 4,
  "num_pre": 2,
  "optimizer": "L-BFGS-B",
  "imaginary": false,
  "safe_hist": true,
  "samples": 20,
  "target_quantum": ["0000", "0110", "1001", "1111"],
  "in_nodes": [0, 1],
  "out_nodes": [2, 3],
  "thresholds": [0.3, 0.1],
  "heralding_out": true,
  "single_emitters": [],
  "amplitudes": [],
  "tries_per_edge": 5,
  "removed_connections": [[0, 3], [1, 2]],
  "seed": null,
  "unicolor": false,
  "number_resolving": true,
  "novac": null,
  "loops": null,
  "topopt": null,
  "dimensions": [],
  "brutal_covers": null,
  "verts": [],
  "anc_detectors": []
}
\end{myverbatim}

\begin{myverbatim}
{
  "description": "Ancilla-assisted qubit-to-ququart multiplexer",
  "foldername": "qubit_to_ququart_mux",
  "bulk_thr": 0.0,
  "edges_tried": 30,
  "ftol": 1e-06,
  "loss_func": "cr",
  "num_anc": 4,
  "num_pre": 1,
  "optimizer": "L-BFGS-B",
  "imaginary": false,
  "safe_hist": true,
  "samples": 10,
  "target_quantum": [
    "10101000",
    "10010100",
    "01100010",
    "01010001"
  ],
  "in_nodes": [0, 1, 2, 3],
  "out_nodes": [4, 5, 6, 7],
  "thresholds": [0.2, 0.1],
  "heralding_out": true,
  "single_emitters": [],
  "amplitudes": [],
  "tries_per_edge": 5,
  "removed_connections": [],
  "number_resolving": true
}
\end{myverbatim}

\begin{myverbatim}
{
  "description": "Measurement‐induced quantum perceptron: two input qubits → one thresholded output qubit",
  "foldername": "quantum_perceptron",
  "bulk_thr": 0.1,
  "edges_tried": 40,
  "ftol": 1e-06,
  "loss_func": "cr",
  "num_anc": 5,
  "num_pre": 1,
  "optimizer": "L-BFGS-B",
  "imaginary": false,
  "safe_hist": true,
  "samples": 100,
  "target_quantum": ["000", "010", "100", "111"],
  "in_nodes": [0, 1],
  "out_nodes": [2],
  "thresholds": [0.3, 0.1],
  "heralding_out": true,
  "single_emitters": [],
  "removed_connections": [[0, 2], [1, 2]],
  "number_resolving": true,
  "tries_per_edge": 5
}
\end{myverbatim}

\begin{myverbatim}
{
  "description": "Entanglement-swapping chain that emulates the Kitaev 1D topological transition: target Bell state on end modes 0 & 1",
  "foldername": "kitaev_swap_chain",
  "bulk_thr": 0.01,
  "edges_tried": 30,
  "ftol": 1e-06,
  "loss_func": "fid",
  "num_anc": 6,
  "num_pre": 1,
  "optimizer": "L-BFGS-B",
  "imaginary": false,
  "safe_hist": true,
  "samples": 50,
  "target_quantum": ["10", "01"],
  "in_nodes": [],
  "out_nodes": [0, 1],
  "thresholds": [0.1, 1],
  "heralding_out": true,
  "tries_per_edge": 5,
  "removed_connections": [[0, 1]],
  "number_resolving": true
}
\end{myverbatim}
\begin{myverbatim}
{
  "description": "Indefinite-order superchannel state generation: superposition of ES→QT and QT→ES branches",
  "foldername": "swap_tp_superchannel",
  "bulk_thr": 0.1,
  "edges_tried": 30,
  "ftol": 1e-06,
  "loss_func": "cr",
  "num_anc": 4,
  "num_pre": 1,
  "optimizer": "L-BFGS-B",
  "imaginary": false,
  "safe_hist": true,
  "samples": 10,
  "target_quantum": [
    "100001",
    "000101"
  ],
  "in_nodes": [],
  "out_nodes": [],
  "thresholds": [0.3, 0.1],
  "single_emitters": [],
  "amplitudes": [1.0, 1.0],
  "tries_per_edge": 5,
  "removed_connections": [
    [0, 2],
    [1, 4],
    [2, 4],
    [3, 0]
  ],
  "number_resolving": true
}
\end{myverbatim}
\newpage
\section{Summaries of the Prompts}
Below, we show summarizations of the prompts used in \AIMandel. The full prompts can be found in our \href{https://github.com/artificial-scientist-lab/ai-mandel}{github repository}.

% One place to control vertical spacing between components
\newcommand{\promptsep}{1.0em}

\subsection{Researcher}
\noindent\textbf{Role, task, and behaviour}. 
The agent proposes novel, Pytheus-compatible targets in quantum optics (quantum networks, protocols, transformations, especially swapping/teleportation, never state generation). It should critically use prior information, avoid ideas too similar to known ones, and aim for targets that are interesting even without implementation and support a broader research programme.

\vspace{\promptsep}

\noindent\textbf{Constraints and capabilities}. 
Targets must be realizable by Pytheus using linear optics, probabilistic pair sources, single-photon emitters, and photon detection (no cavities/resonances), and should respect the explicit capabilities and limitations given in the prompt.

\vspace{\promptsep}

\noindent\textbf{Inputs}. 
The agent is given a concise description of Pytheus' capabilities and limitations, a few random arXiv paper abstracts in quantum physics, a list of previously explored target experiments, several example Pytheus configurations, and a required pair of scientific concepts that must be meaningfully combined.

\vspace{\promptsep}

\noindent\textbf{Available actions and output format}. 
The agent can either search literature with \texttt{arxiv} or output a final proposal. It must respond in exactly one of the following forms:
\begin{quote}
\begin{verbatim}
Thought: (reflect on your progress and decide what to do next)
Action: arxiv
Action Input: (the input string to the action)
\end{verbatim}
\end{quote}
or
\begin{quote}
\begin{verbatim}
Thought: (use a 5-step internal template to brainstorm, check novelty and
    feasibility, and justify the final choice)
Action: final answer
Action Input: (well-formed target description for the Expert, in the style of the
    example configurations, including a visualization of the network structure)
\end{verbatim}
\end{quote}

\vspace{2\promptsep}

\newpage
\subsection{Novelty Supervisor}
\noindent\textbf{Role and task}. 
The agent acts as a critical supervisor focused on \emph{novelty}, evaluating Researcher-proposed targets (foundational experiments, quantum networks, or protocols) in quantum optics and deciding whether they merit further study.

\vspace{\promptsep}

\noindent\textbf{Novelty criteria}. 
Ideas must go beyond simple state-generation experiments or trivial extensions of known work, be clearly distinct from prior examples, and plausibly reach the level of a selective journal such as PRL.

\vspace{\promptsep}

\noindent\textbf{Inputs}. 
Novelty is judged against two lists: human-explored experiments (a catalogue of previously implemented Pytheus targets) and ensemble-proposed ideas (a collection of abstracts from other agent-generated proposals); the agent should identify and point out close parallels where possible.

\vspace{\promptsep}

\noindent\textbf{Available actions and output format}. 
The agent can \texttt{accept} or \texttt{reject} the idea. For rejections, it must respond as:
\begin{quote}
\begin{verbatim}
Thought: (reflect on the idea given by the researcher and whether it can be
    considered novel.)
Action: (the action name, should be 'reject')
Action Input: (express your reasoning for rejecting the idea and what has to
    be improved.)
\end{verbatim}
\end{quote}
Accepted ideas use the same structure with \texttt{Action: accept} and a concise justification, and are passed on as promising candidates.

\newpage
\subsection{Judge}
\noindent\textbf{Role and task}. 
The agent is an expert quantum physicist who evaluates a proposed research idea (described in natural language) for novelty and feasibility within the Pytheus framework, using a detailed description of Pytheus (its architecture, capabilities, and limitations) provided in the prompt.

\vspace{\promptsep}

\noindent\textbf{Evaluation criteria}. 
It checks whether the physics is coherent, whether a Pytheus-compatible photonic experiment could realistically implement the idea, and whether the idea is sufficiently novel, rejecting any proposal that treats \emph{state generation} itself as the target.

\vspace{\promptsep}

\noindent\textbf{Available actions}. 
The agent can \texttt{accept} an idea (if sound, feasible, and novel enough) or \texttt{reject} it (if unsound, infeasible, unoriginal, or forbidden due to state generation).

\vspace{\promptsep}

\noindent\textbf{Output format}. 
For rejections, the agent must respond as:
\begin{quote}
\begin{verbatim}
Thought: (explain your understanding of the proposed experiment and your
    reasoning on its feasibility and novelty.)
Action: (the action name, should be 'reject')
Action Input: (Provide your final feedback as a self-contained two to three
    sentence response (not referencing your previous statements). Do not
    suggest changing the idea to one that has state generation as a target.)
\end{verbatim}
\end{quote}
Accepted ideas use the same structure with \texttt{Action: accept} and a positive, self-contained two- to three-sentence justification; these are passed along for further consideration.

\vspace{2\promptsep}

\newpage
\subsection{Expert}
\noindent\textbf{Role and overall task}. 
The agent takes a conceptual idea and converts it into a concrete Pytheus configuration, then calls the \texttt{pytheus} tool. The configuration must faithfully represent the proposed quantum optics experiment.

\vspace{\promptsep}

\noindent\textbf{Configuration file structure}. 
The configuration is a flat JSON object with: (i) a short experiment description, (ii) a precise target specification (state/measurement/transformation/protocol), and (iii) suitable optimization and resource parameters, following the structure visible in the provided example configurations.

\vspace{\promptsep}

\noindent\textbf{Inputs}. 
The agent receives a concise description of Pytheus' capabilities and limitations, a set of example Pytheus configurations as references, a short abstract summarizing the idea, and the full natural-language description of the proposed experiment.

\vspace{\promptsep}

\noindent\textbf{Action and output format}. 
The only action is \texttt{pytheus}, and the agent must reply in:
\begin{quote}
\begin{verbatim}
Thought: (reflect on your progress and decide what to do next)
Action: (the action name, should be 'pytheus')
Action Input: (the input string to the action, which is a string in full JSON
    syntax, similar to the examples above, only config, no explanation needed,
    single level)
\end{verbatim}
\end{quote}

\vspace{2\promptsep}

\end{document}